\documentclass[runningheads]{llncs}
\usepackage{graphicx}
\usepackage{subcaption}
\usepackage{float}
\usepackage{amsmath,amsfonts,bm}
\usepackage{multirow}
\usepackage{booktabs}

\usepackage{hyperref}
\hypersetup{hypertex=true,
colorlinks=true,
linkcolor=blue,
anchorcolor=blue,
citecolor=blue}

\begin{document}
%
\title{Multi-Scale Correlation-Aware Transformer for Maritime Vessel Re-Identification}
\titlerunning{MCFormer for Maritime Vessel Re-ID}
%
\author{Yunhe Liu}
\authorrunning{Yunhe Liu}
\institute{Tsinghua Shenzhen International Graduate School, Tsinghua University, Shenzhen 518055, China \\
\email{liuyunhe23@mails.tsinghua.edu.cn}}
%
\maketitle              
\begin{abstract}
Maritime vessel re-identification (Re-ID) plays a crucial role in advancing maritime monitoring and intelligent situational awareness systems. However, some existing vessel Re-ID methods are directly adapted from pedestrian-focused algorithms, making them ill-suited for mitigating the unique problems present in vessel images, particularly the greater intra-identity variations and more severe missing of local parts, which lead to the emergence of outlier samples within the same identity. To address these challenges, we propose the Multi-scale Correlation-aware Transformer Network (MCFormer), which explicitly models multi-scale correlations across the entire input set to suppress the adverse effects of outlier samples with intra-identity variations or local missing, incorporating two novel modules, the Global Correlation Module (GCM), and the Local Correlation Module (LCM). Specifically, GCM constructs a global similarity affinity matrix across all input images to model global correlations through feature aggregation based on inter-image consistency, rather than solely learning features from individual images as in most existing approaches. Simultaneously, LCM mines and aligns local features of positive samples with contextual similarity to extract local correlations by maintaining a dynamic memory bank, effectively compensating for missing or occluded regions in individual images. To further enhance feature robustness, MCFormer integrates global and local features that have been respectively correlated across multiple scales, effectively capturing latent relationships among image features. Experiments on three benchmarks demonstrate that MCFormer achieves state-of-the-art performance.

\end{abstract}
\section{Introduction}
\subsubsection{Motivation.}
\begin{figure}[H]
    \centering
    \begin{subfigure}[a]{0.65\textwidth}
        \centering
        \includegraphics[width=\textwidth]{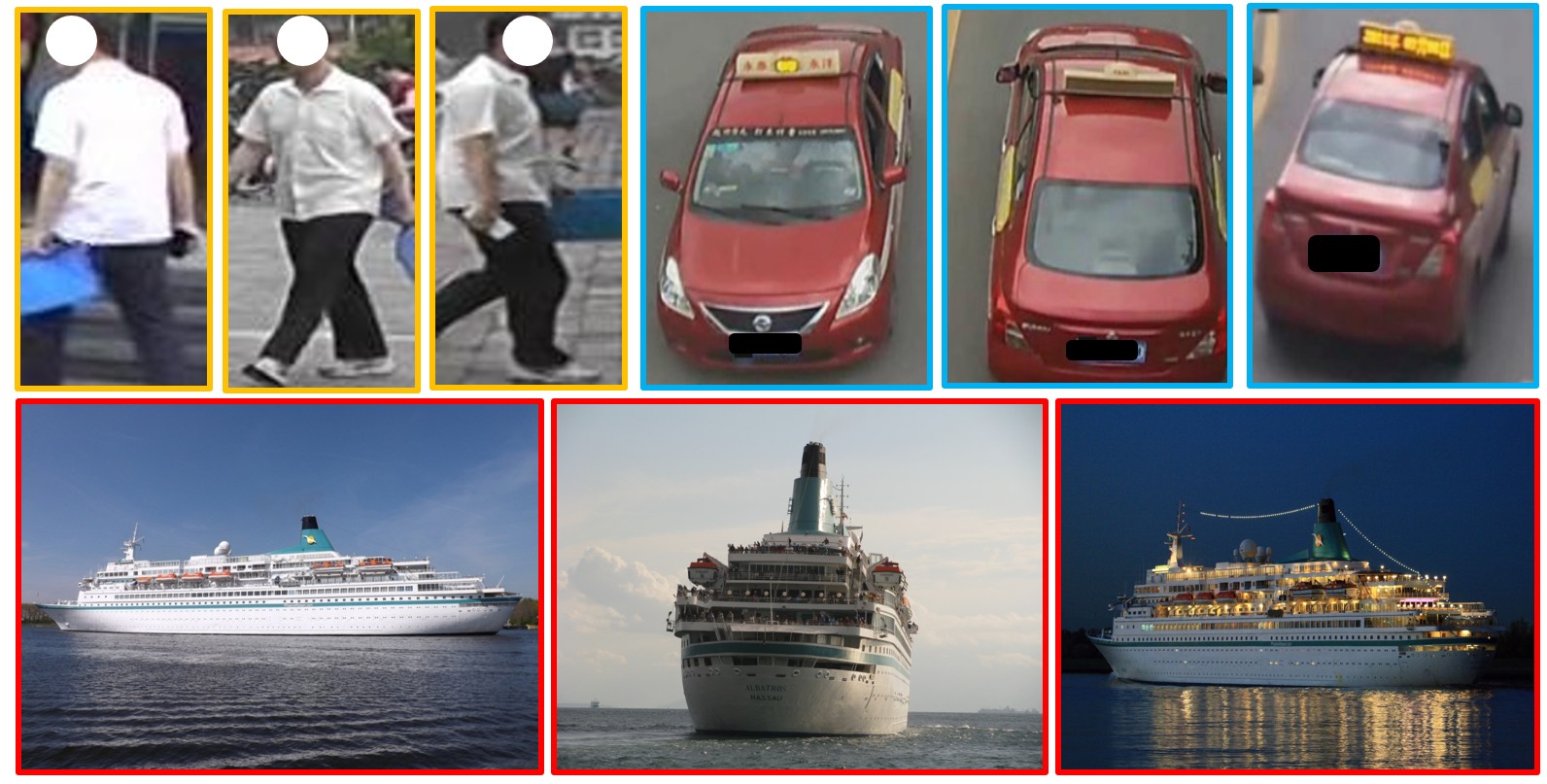}
        \caption{} 
        \label{fig1a}
    \end{subfigure}
    \begin{subfigure}[b]{0.65\textwidth}
        \centering
        \includegraphics[width=\textwidth]{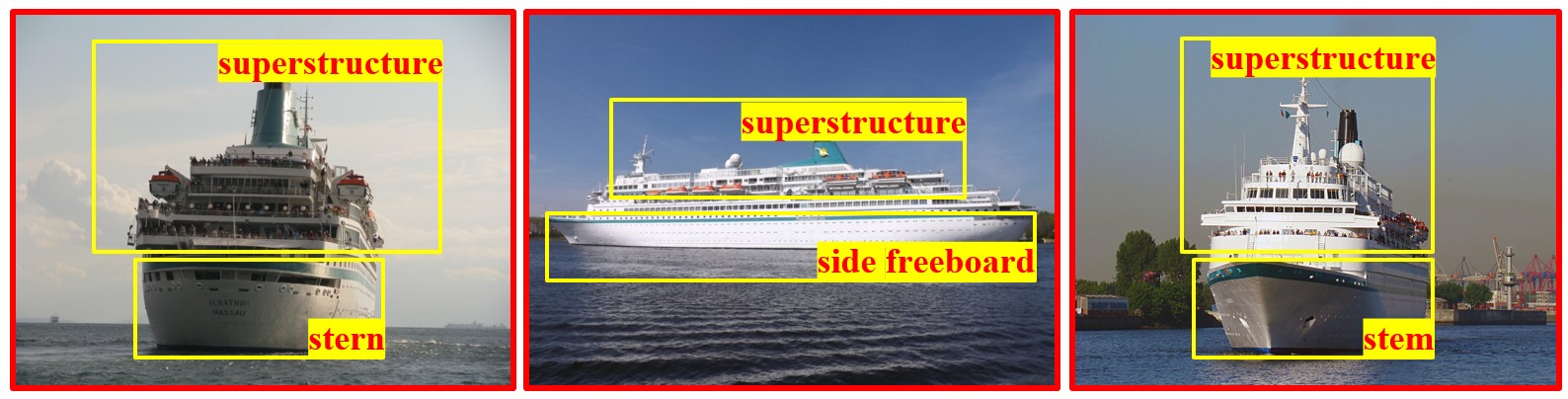}
        \caption{} 
        \label{fig1b}
    \end{subfigure}

  \caption{(a) Comparison of intra-identity variations in person/vehicle and vessel images, where the image frames in the borders of same color denote that the corresponding images share the same person/vehicle/vessel identity. (b) Illustration of vessel part annotations and local missing in different viewing conditions. Each image of vessel exhibits partial occlusion or missing regions due to varying viewpoint and distance.}
  \label{fig1}
\end{figure}

Identifying and retrieving specific vessel targets from extensive maritime surveillance data is crucial for developing intelligent maritime monitoring and situational awareness systems. Vessel re-identification (Re-ID) techniques play a key role in this endeavor, aiming to retrieve images of the same identity across diverse imaging conditions, platforms, and temporal intervals based on a given query vessel image~\cite{ref1,ref2,ref3}. Compared to rapidly advancing tasks of pedestrian and vehicle Re-ID, vessel Re-ID faces unique challenges, particularly the greater intra-identity variations and more severe missing of local parts. As illustrated in Fig.~\ref{fig1}.(a), pedestrian and vehicle images typically maintain consistent visual appearances despite viewpoint changes, while vessel images exhibit significant variability in viewpoint, structure, and illumination. Furthermore, as shown in Fig.~\ref{fig1}.(b), each vessel can be decomposed into several distinct structural components, while each vessel image may only contain a subset of these components, leading to varying degrees of local missing. However, some existing vessel Re-ID methods are directly adapted from algorithms focused on person, making them ill-suited for addressing the unique problems inherent in vessel images, which lead to the emergence of outlier samples within the same identity. These approaches typically extract discriminative features from individual images using attention mechanisms, local feature extraction, or domain generalization methods, yet they often neglect inter-image correlations that can provide complementary contextual information. 

\subsubsection{Our Approach.}
To this end, we propose the Multi-Scale Correlation-Aware Transformer Network (MCFormer), which explicitly models inter-image correlations at both global and local scales, as illustrated in Fig.~\ref{fig2}. Based on the powerful long-range dependency modeling capability of Transformer architectures, MCFormer effectively captures complementary information across images to suppress the adverse effects of outlier samples with intra-identity variations or local missing. Specifically, we propose a Global Correlation Module (GCM) that constructs a similarity affinity matrix across all input images to model global correlations through feature aggregation. GCM projects high-dimensional global features into a lower-dimensional space via a small set of randomly sampled landmarks to reduce the computations in the affinity matrix. This relation modeling between images suppresses high intra-identity variations and leads to more robust features. In parallel, we introduce a Local Correlation Module (LCM) that mines and aligns local features of positive samples with contextual similarity to extract local correlations. LCM maintains a dynamic memory bank of local features and identify the nearest top-k neighbors as positive samples. By applying an optimized center loss, LCM minimizes the distance between these local features to compensate for missing of local parts. Finally, the globally and locally correlated features are fused through a multi-scale channel attention mechanism, which adaptively emphasizes informative channels across scales. This fusion enhances the ability to extract both robust and generalizable representations under diverse conditions of our model. 

\textbf{Contributions.} Our main contributions of this paper are summarized as:
\begin{quote}
    (1) We propose a Global Correlation Module to alleviate large intra-identity variations by modeling holistic global feature relationships across all input images.
    
    (2) We introduce a Local Correlation Module to compensate for the missing of local parts through neighboring positive samples with contextual similarity.
    
    (3) We integrate the globally and locally correlated features, using a multi-scale channel attention mechanism, resulting in a novel Multi-Scale Correlation-Aware Transformer Network. 
    
    (4) Extensive experiments demonstrate that MCFormer achieves state-of-the-art performance on vessel Re-ID datasets, validating its effectiveness in capturing latent inter-image relationships among features.
\end{quote}
\section{Related Work}
\subsection{Vessel Re-Identification}
Compared to extensive studies on pedestrian and vehicle Re-ID, vessel Re-ID has received significantly less attention, primarily due to the limited availability of publicly accessible vessel datasets at the instance level~\cite{ref4,ref5,ref6}. However, this situation has gradually improved in recent years, with more vessel Re-ID algorithms and datasets being publicly released. Spagnolo et al.~\cite{ref9} introduced the first publicly available dataset for vessel re-identification, BoatRe-ID, which consists of 107 vessel identities and 5,523 images. They also proposed a baseline for vessel Re-ID based on discriminative CNN embedding. Zeng et al.~\cite{ref10} proposed a transfer learning-based approach to mitigate the impact of vessel oscillation. They also constructed the Warships-ReID dataset, which contains 163 vessel identities and 4,780 images. Zhang et al.~\cite{ref11} introduced a multi-level contrastive learning framework for vessel Re-ID, trained with a specifically designed intra-batch cluster-level contrastive loss along with an instance-level loss. They also developed the VesselReID dataset, which includes 1,248 vessel identities and 30,587 images. These studies have laid a solid foundation for the development of vessel Re-ID. However, as discussed before, most of these methods focus exclusively on extracting features from individual images, while neglecting the inter-image correlations that can provide valuable complementary information. This oversight often results in suboptimal performance.
\subsection{Correlation-Related Re-ID}
Transformer architectures have shown strong capabilities in modeling long-range dependencies in computer vision tasks, thanks to their multi-head attention mechanism. Recent studies have begun to explore the use of Transformers for correlation-aware Re-ID, particularly in pedestrian scenarios~\cite{ref13,ref14,ref15}. However, most of these works are limited to modeling correlations within small image batches. He et al.~\cite{ref17} were among the first to adopt a pure Transformer architecture for person Re-ID research. Ni et al.~\cite{ref18} proposed a locally correlated Transformer for generalizable person Re-ID, aiming to alleviate the impact of domain-specific biases. Wang et al.~\cite{ref19} introduced a robust person Re-ID model using the Neighbor Transformer, which applies sparse attention to relevant neighbors instead of all. Zhang et al.~\cite{ref20} proposed a Local Graph Attention Network, which models both inter-local and intra-local relations within a complete local graph. Zhu et al.~\cite{ref21} introduced Dual Cross-Attention, which reduces misleading attentions and diffuses attention responses to discover more complementary features for recognition. Unlike the aforementioned methods, our proposed approach simultaneously models global correlations from complete input images and local correlations from neighboring samples. This dual-level correlation modeling is particularly suited to the challenges of vessel Re-ID, where both large intra-identity variation and local parts missing are prevalent. As a result, our method offers improved robustness capabilities over prior approaches.
\section{Method}
\begin{figure}[H]
\centering
\includegraphics[width=\textwidth]{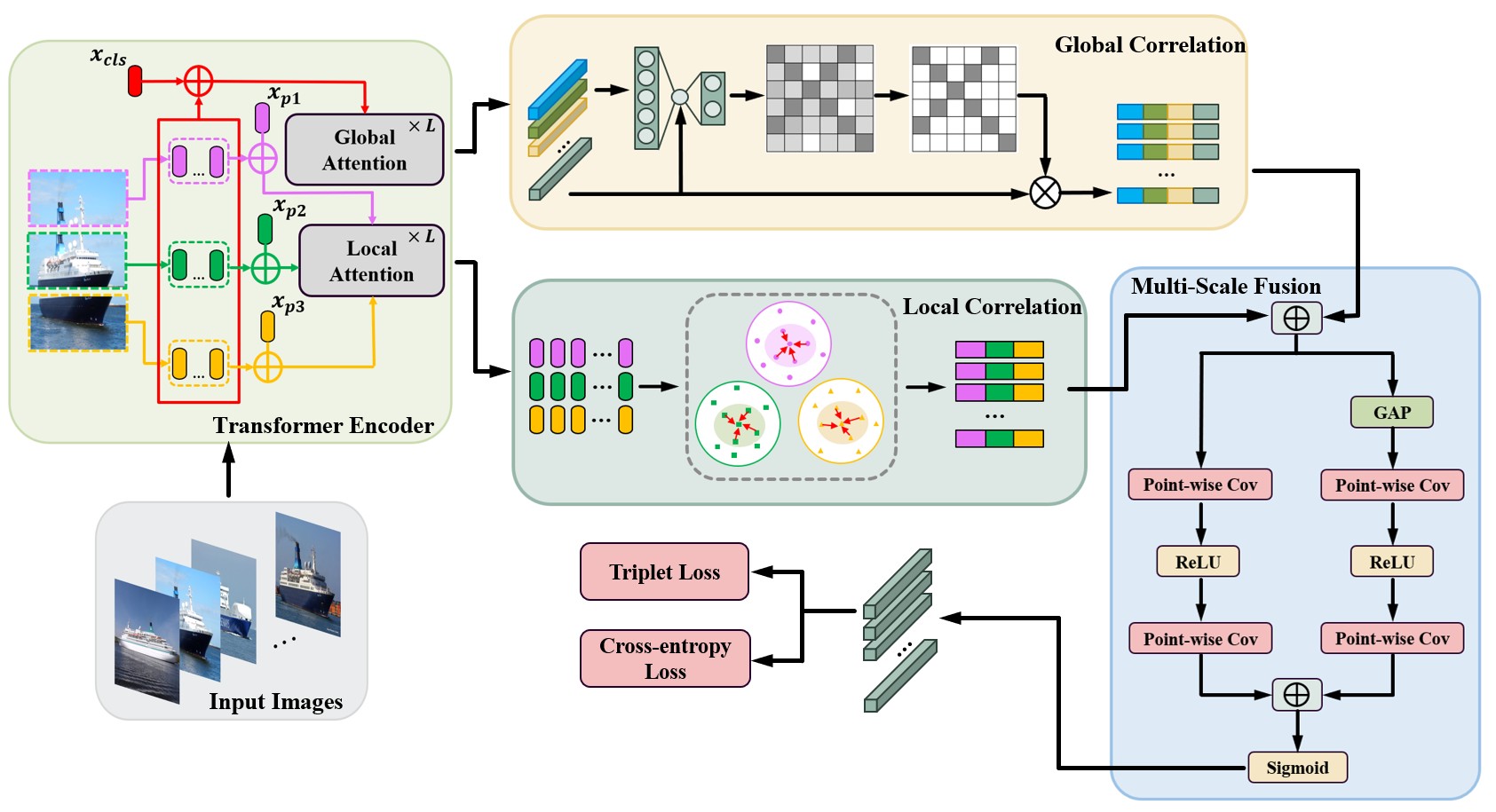}
\caption{An illustration of our proposed MCFormer. The input images is first fed into the Transformer Encoder, which extracts global features using the class token $x_{\mathrm{cls}}$ and all image tokens, and local features using the part tokens $x_{p_i}$ and the corresponding regional image tokens. The global features are processed by the GCM, where they are first mapped from a high-dimensional space to a lower-dimensional space using a small set of randomly sampled landmarks. A global affinity matrix is then computed and sparsified, followed by a weighted aggregation to obtain the correlated global representations. The local features are processed by the LCM, where positive samples are identified and a clustering loss is applied to pull neighboring positive samples closer together. The correlated local representation is then obtained through concatenation and dimensionality reduction. Finally, the correlated global and local features are fused via a multi-scale channel attention module, resulting in a robust representation used for vessel Re-ID retrieval.} \label{fig2}
\end{figure}

In this section, we present the detailed structure of the proposed MCFormer, and an overview of our model designed for vessel Re-ID is presented in Fig.~\ref{fig2}.
\subsection{Transformer Encoder}
Our Transformer encoder consists of global attention layer and local attention layer, responsible for extracting global and local features respectively. We define the input dataset as \(\bigl\{\bm{x}^i, y^i\bigr\}_{i=1}^D\), where \(\bm{x}^i\) represents the \(i\)-th image and \(y^i\) is its identity label.  Each input image \(x\) is divided into \(N\) non-overlapping patches by
\noindent
the Transformer’s embedding module, yielding the “image tokens”, denoted as \(\{\,x_i \mid i=1,2,\dots,N\}.\) Additionally, we introduce a learnable “class token” \(x_{\mathrm{cls}}\) and three “part tokens” \(\{x_{p_i}\mid i=1,2,3\}\), which are concatenated with all image tokens. The Transformer encoder input is then formulated as:
\begin{equation}X = [\,x_{\mathrm{cls}},\,x_{p_1},\,x_{p_2},\,x_{p_3},\,x_1,\dots,x_N\,] + P,\end{equation}
where \(P\) is the positional embedding. \(
Q = \bigl[q_{\mathrm{cls}},\,q_{p_1},\,q_{p_2},\,q_{p_3},\,q_1,\dots,q_N\bigr],\,
K = \bigl[k_{\mathrm{cls}},\,k_{p_1},\,k_{p_2},\,k_{p_3},\,k_1,\dots,k_N\bigr],\,
V = \bigl[v_{\mathrm{cls}},\,v_{p_1},\,v_{p_2},\,v_{p_3},\,v_1,\dots,v_N\bigr]
\) are projections in the attention mechanism. To extract global features, we compute self-attention using the class token and all image tokens:
\begin{equation}
  \mathrm{Attention}(Q_{\mathrm{cls}},K_{\mathrm{cls}},V_{\mathrm{cls}})
    = \mathrm{Softmax}\Bigl(\frac{Q_{\mathrm{cls}}\,K_{\mathrm{cls}}^T}{\sqrt{d}}\Bigr)\,V_{\mathrm{cls}},
\end{equation}
where \(d\) is the embedding dimension, and \(
  Q_{\mathrm{cls}}=[\,q_{\mathrm{cls}},q_1,\dots,q_N\,],\,
  K_{\mathrm{cls}}=[\,k_{\mathrm{cls}},k_1\\,\dots,k_N\,],\,
  V_{\mathrm{cls}}=[\,v_{\mathrm{cls}},v_1,\dots,v_N\,].
\) Repeating for \(L\) layers, we obtain the global feature set \(\{\bm{g}^i\}_{i=1}^D\). To extract local features, each part token \(x_{p_i}\) attends only to its corresponding patch subset:
\begin{equation}
  \mathrm{Attention}(Q_{p_i},K_{p_i},V_{p_i})
    = \mathrm{Softmax}\Bigl(\frac{Q_{p_i}\,K_{p_i}^T}{\sqrt{d}}\Bigr)\,V_{p_i},
\end{equation}
where \(Q_{p_i}=[\,q_{p_i},q_{k_i},\dots,q_{k_{i+m}}\,],\,
K_{p_i}=[\,k_{p_i},k_{k_i},\dots,k_{k_{i+m}}\,],\,
V_{p_i}=[\,v_{p_i},v_{k_i},\dots\\,v_{k_{i+m}}],\) and \(\{k_i,\dots,k_{i+m}\}\) are the indices of image tokens in region \(i\). Repeating for \(L\) layers, we obtain the local feature set \(\{\bm{l}_1^i,\bm{l}_2^i,\bm{l}_3^i\}_{i=1}^D\).

\subsection{Global Correlation Module}
We aim to model the correlation among global features of the entire input dataset. To explicitly describe this correlation, we introduce a function to obtain the correlated global representation \(\bm{u}^i\):
\begin{equation}\label{5}
  \bm{u}^i
    = h\bigl(\bm{g}^i,\{\bm{g}^j\}_{j=1}^D\bigr)
    = \sum_{j=1}^D \bm{w}_{ij}\,\bm{g}^j,
    \quad
    \sum_j \bm{w}_{ij}=1,
\end{equation}
where \(\bm{w}_{ij}\) is a learnable weight of the correlation between \(\bm{g}^i\) and \(\bm{g}^j\). Following the self-attention formulation, equation~\eqref{5} can be rewritten as:
\begin{equation}\label{eq:6}
  \bm{u}^i \;=\; \sum_{j} s(\bm{A})_{ij}\,\phi_v\bigl(\bm{g}^j\bigr),
\end{equation}
where \(\bm{A}\in\mathbb{R}^{N\times N}\) is the affinity matrix capturing pairwise correlations between global features \(\bm{g}^i\) and \(\bm{g}^j\). \(s(\cdot)\) denotes the softmax over each row of \(\bm{A}\), and \(\phi_v(\cdot)\) is a learnable linear projection. The affinity matrix \(\bm{A}\) itself is computed as
\begin{equation}\label{eq:7}
  \bm{A}_{ij}
    = \frac{K\bigl(\phi_q(\bm{g}^i),\,\phi_k(\bm{g}^j)\bigr)}{\sqrt{d}}
    = \frac{\bm{q}_i\,\bm{k}_j^{T}}{\sqrt{d}},
\end{equation}
where \(\phi_q,\phi_k:\mathbb{R}^d\to\mathbb{R}^d\) are two linear mappings that transform the input representation \(\bm{g}\in\mathbb{R}^{N\times d}\) into query and key matrices \(\bm{q}, \bm{k}\in\mathbb{R}^{N\times d}\) , and \(K(\cdot,\cdot)\) is the inner‐product. To reduce complexity, we randomly sample \(\ell\ll d\) landmark features from \(\bm{g}\), denoted \(\bm{g}_\ell\in\mathbb{R}^{\ell\times d}\), to serve as landmark proxies. The query and key matrices are then computed from these landmarks, allowing dimensionality reduction. Consequently, equation~\eqref{eq:7} can be reformulated as:
\begin{equation}\label{eq:8}
  \bm{A}'_{ij}
    = \frac{(\bm{q}\,\bm{k}_\ell^T)_i\,(\bm{k}\,\bm{q}_\ell^T)_j^{T}}{\sqrt{d}}
    = \frac{\bm{q}'_i\,(\bm{k}'_j)^{T}}{\sqrt{d}},
\end{equation}
where \(\bm{q}',\bm{k}'\in\mathbb{R}^{N\times \ell}\). Since \(\ell\ll d\), this reduces the computational complexity of affinity matrix multiplication from \(O(N^2 d)\) to \(O(N^2 \ell)\). During the softmax normalization of \(\bm{A}'\), we apply a reciprocal‐neighbor mask to sparsify the attention weights for computational efficiency. The sparse masked attention weights become
\begin{equation}\label{eq:10}
  s(\bm{A}')_{ij}
    = \frac{M_{ij}\,\exp\bigl(-\bm{A}'_{ij}\bigr)}
           {\sum_{k} M_{ik}\,\exp\bigl(-\bm{A}'_{ik}\bigr)},
           \quad
    M_{ij} =
  \begin{cases}
    1, \,\, j\in\mathrm{tk}(\bm{A}'_{i,:}) , i\in\mathrm{tk}(\bm{A}'_{:,j})\\
    0, \,\, \text{otherwise}
  \end{cases}
\end{equation}
where \(\mathrm{tk}(\cdot)\) returns the indices of the top-k entries. By applying these sparse attention weights to equation~\eqref{eq:6}, we obtain the final correlated global representation \(\{\bm{u}^i\}_{i=1}^D\), which integrates information from all input images.

\subsection{Local Correlation Module}
We maintain a momentum‐updated memory bank \(\{w_{p_i}^j\}\) for each local feature:
\begin{equation}\label{eq:11}
  w_{p_i}^j =
  \begin{cases}
    \bm{l}_i^j,                      & t = 0,\\
    (1 - m)\,w_{p_i}^j + m\,\bm{l}_i^j, & t > 0,
  \end{cases}
\end{equation}
where \(t\) is the training epoch, \(m\) is the momentum coefficient, and \(j=1,\dots,D\). For each local feature \(\bm{l}_i^j\), we compare it against the entire memory bank \(\{w_{p_i}^n\}_{n=1}^D\) and select the \(k\) most similar as the positive set \(P_{p_i}^j\).  We then define the clustering loss to minimize the distance between \(\bm{l}_i^j\) and its positive samples:
\begin{equation}\label{eq:12}
L_{p_i}^j
= -\log
  \frac{\displaystyle\sum_{w_{p_i}^m\in P_{p_i}^j}
        \exp\bigl(\tfrac{\bm{l}_i^j\cdot w_{p_i}^m}{\tau}\bigr)}
       {\displaystyle\sum_{n=1}^D
        \exp\bigl(\tfrac{\bm{l}_i^j\cdot w_{p_i}^n}{\tau}\bigr)},
\end{equation}
where \(\tau\) is the temperature.  Minimizing \(L_{p_i}^j\) pulls positives closer in feature space and pushes negatives away.  Finally, the three local features are concatenated and reduced by point‐wise convolution to yield the deeply correlated local representation \(\{\bm{v}^i\}_{i=1}^D\).

\subsection{Multi-Scale Fusion}
After obtaining the correlated global and local features respectively, we fuse them via multi‐scale channel attention. Given a global feature \(\bm{u}\) and local feature \(\bm{v}\in\mathbb{R}^{C\times H\times W}\), the fused representation is
\begin{equation}\label{eq:13}
z \;=\; m(\bm{u}\oplus \bm{v})\,\otimes\,\bm{u} \;+\;\bigl(1 - m(\bm{u}\oplus \bm{v})\bigr)\,\otimes\,\bm{v},
\end{equation}
where \(\oplus\) is broadcast addition, \(\otimes\) elementwise multiplication, and
\begin{equation}\label{eq:14}
m(\bm{x})
= C_2\bigl(\delta\bigl(C_1(\bm{x})\bigr)\bigr)
  \;\oplus\;
  \sigma\bigl(C_2\bigl(\delta\bigl(C_1(g(\bm{x}))\bigr)\bigr)\bigr).
\end{equation}
Here \(g(\bm{x})=\tfrac{1}{H\times W}\sum_{i=1}^H\sum_{j=1}^W X_{:,i,j}\) is global average pooling, \(\delta\) (ReLU) and \(\sigma\) (Sigmoid) are activations, and \(C_1,C_2\) are \(1\times1\) convolutions of size \(\tfrac{C}{r}\times C\) and \(C\times\tfrac{C}{r}\), respectively. This fusion reduces the impact of scale variation and partial occlusion effectively.

\section{Experiments}
\subsection{Datasets, Evaluation Protocols and Implementation}
\subsubsection{Datasets.}
Experiments were conducted on three publicly available vessel re-identification datasets: VesselReID~\cite{ref11}, Warships-ReID~\cite{ref10}, and BoatRe-ID~\cite{ref9}. Each vessel in VesselReID and Warships-ReID is labeled with its International Maritime Organization number, and images were collected from the platforms of website. BoatRe-ID is used primarily for supplementary evaluation due to its relatively small scale. All of the datasets mentioned above contain multiple images for each identity collected from various conditions and cameras. Detailed statistics of each dataset are summarized in Table~\ref{tab1}.
\begin{table}
\centering
\caption{Statistics of vessel ReID datasets.}\label{tab1}
\begin{tabular*}{0.6\textwidth}{@{\extracolsep{\fill}}c c c}
\hline
\bfseries Dataset & \bfseries IDs & \bfseries Images\\
\hline
VesselReID &  1248 & 30587\\
Warships-Reid & 163 & 4780\\
BoatRe-ID & 107 & 5523\\
\hline
\end{tabular*}
\end{table}

\textbf{Evaluation Protocols.}
We follow the standard evaluation protocols for vessel Re-ID, using two widely adopted evaluation metrics: mean Average Precision (mAP) and Cumulative Matching Characteristics (CMC) at Rank-1 (R1), Rank-5 (R5), Rank-10 (R10). The CMC curve measures the probability that a correct match appears within the top-k ranked results, while mAP evaluates the area under the precision-recall curve, offering a more comprehensive assessment of overall retrieval performance.

\textbf{Implementation.}
The proposed MCFormer is built upon a Vision Transformer (ViT-B/16)~\cite{ref22} pre-trained on ImageNet, as the encoder backbone, with a \({stride=16}\). Input images are resized to 256 × 128. For data augmentation, we apply random horizontal flipping and local grayscale transformations. The model is optimized using the SGD optimizer with weight decay set to \({10}^{-4}\). The learning rate is initially set to 0 and linearly increased to \({10}^{-3}\) during the first 10 epochs, then decays during the following 90 epochs. The training process consists of three stages, each trained for 100 epochs: (1) Use ViT-B/16 to extract global and local features, with a batch size of 128. (2) Perform global and local correlation using the MCFormer framework, with the batch size set to the size of the full training set. (3) Apply multi-scale feature fusion, with a batch size of 128. All experiments were conducted on a GeForce RTX 4090 GPU using PyTorch.

\subsection{Comparison with State-of-the-art Methods}
We compare MCFormer with SOTA methods on the three vessel Re-ID benchmarks. The experimental results show that the proposed MCFormer outperforms other SOTA methods in both Rank-n and mAP performance metrics.

\begin{table}
  \centering
  \caption{Comparison with SOTA Methods.}\label{tab:comparison}
  \begin{tabular*}{\textwidth}{@{\extracolsep{\fill}}l
                               cccc|cccc|cc}
    \toprule
    \multirow{2}{*}{\bfseries Methods}
      & \multicolumn{4}{c|}{\bfseries VesselReID}
      & \multicolumn{4}{c|}{\bfseries Warships-ReID}
      & \multicolumn{2}{c}{\bfseries BoatRe-ID} \\
    & \bfseries R1 & \bfseries R5 & \bfseries R10 & \bfseries mAP
    & \bfseries R1 & \bfseries R5 & \bfseries R10 & \bfseries mAP
    & \bfseries R1 & \bfseries mAP \\
    \midrule
    BNN~\cite{ref23}         & 63.8 & 85.3 & 90.9 & 50.7  & 87.5 & 95.8 & 95.8 & 70.3  & 96.5 & 89.6 \\
    PASS~\cite{ref18}         & 66.2 & 85.7 & 90.9 & 53.9  & 91.5 & 96.3 & 96.6 & 75.2  & 97.9 & 90.6 \\
    MCL~\cite{ref11}          & 63.9 & 82.1 & 88.9 & 45.3  & 86.1 & 91.1 & 91.9 & 66.9  & 85.0 & 63.6 \\
    Trans-Reid~\cite{ref17}   & 68.2 & 86.3 & 91.1 & 58.7  & 92.1 & 97.4 & 97.4 & 77.1  & 98.2 & 92.1 \\
    PFD-Net~\cite{ref24}      & 66.1 & 84.9 & 90.2 & 49.2  & 90.5 & 96.5 & 96.5 & 73.9  & 96.3 & 90.3 \\
    AP-Net~\cite{ref25}       & 62.6 & 83.3 & 89.6 & 50.1  & 82.9 & 97.1 & 97.1 & 53.2  & 95.9 & 89.2 \\
    Tran-Aligned~\cite{ref10} & 64.3 & 82.8 & 89.4 & 51.8  & 88.6 & 97.1 & 97.1 & 62.5  & 97.5 & 91.1 \\
    \textbf{MCFormer} & \textbf{72.8} & \textbf{88.9} & \textbf{93.1} & \textbf{63.4}
                      & \textbf{96.1} & \textbf{98.1} & \textbf{98.1} & \textbf{88.4}
                      & \textbf{98.8} & \textbf{93.5} \\
    \bottomrule
  \end{tabular*}
\end{table}

\noindent
\textbf{Results on VesselReID.}
As shown in Table~\ref{tab:comparison}, MCFormer surpasses the strongest baseline (Trans-Reid) by +4.6\% in terms of Rank-1, and improvements of +2.6\% and +2.0\% are observed in Rank-5 and Rank-10. Notably, MCFormer reaches an mAP of 63.4\%, which is +4.7\% higher than Trans-Reid and +12.7\% higher than early methods like BNN. This demonstrates the effectiveness of our multi-scale correlation modeling, which captures both global structural alignment and local complementary features, is crucial in datasets like VesselReID where large viewpoint and appearance variations are common.

\textbf{Results on Warships-Reid.}
Table~\ref{tab:comparison} shows that MCFormer also achieves superior performance on Warships-ReID. While Rank-n gains are modest (e.g., +0.7\% in Rank-5 and Rank-10), the large margin in mAP reflects MCFormer’s ability to retrieve true matches with higher confidence across the ranking spectrum. This is particularly valuable in Warships-ReID, which contains vessels of similar appearance under challenging conditions (e.g., waves, and simulated motion), where discriminative feature aggregation plays a key role.

\textbf{Results on BoatRe-ID.}
Compared to the strong baseline Trans-Reid and Tran-Aligned, MCFormer shows consistent improvements across all ranks. The higher mAP indicates that MCFormer not only retrieves correct matches at top positions but also maintains better ranking quality overall. This demonstrates the effectiveness of our global-local correlation modeling and multi-scale fusion strategy, even on relatively small-scale datasets like BoatRe-ID. 
\begin{figure}
\centering
\includegraphics[width=\textwidth]{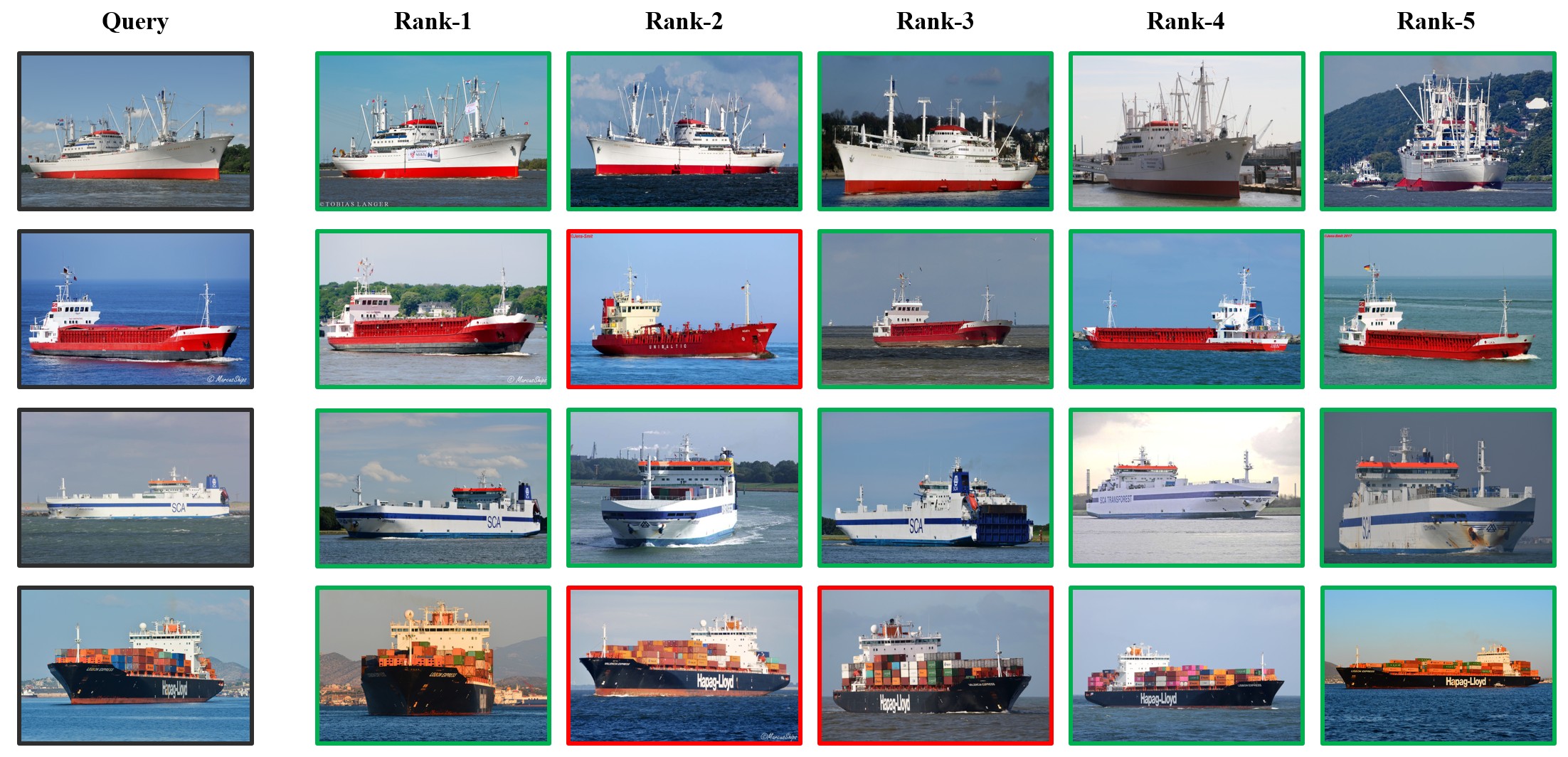}
\caption{Visualization of the ranking results on the VesselReID test set. Green and red boxes indicate correct and incorrect retrieval results respectively.} \label{ranking results}
\end{figure}

As shown in Fig.~\ref{ranking results}, we present the re-identification results on the testset of VesselReID using MCFormer. MCFormer successfully retrieves most of the correct matches within the top-ranked results, as indicated by the green bounding boxes. The model shows a strong ability to identify visually similar images belonging to the same identity, achieving high Rank-n accuracy. However, some matches in lower ranks are occasionally misclassified (red boxes), which reflects a lack of strong feature centralization across all samples of the same identity. This phenomenon is consistent with the observed discrepancy between high Rank-n scores and relatively lower mAP values. We attribute this to the high visual similarity between vessel images of different identities within the same category, which makes it difficult for the model to learn tightly clustered feature distributions. This is the key issue that we need to address in our future work.
\subsection{Ablation Study}
\textbf{Effectiveness of main components of MCFormer.}
To ensure that both GCM and LCM contribute to our model, we conduct a ablation study on VesselReID and Warships-Reid. Table~\ref{tab:components} presents the performance comparison of the following models on the two datasets: (1) Baseline, standard ViT-B/16. (2) Model trained with only GCM. (3) Model trained with only LCM. (4) Complete MCFormer. On the VesselReID dataset, incorporating GCM alone improves Rank-1/mAP by +3.5\%/+7.7\% over the baseline. Similarly, on Warships-Reid, the GCM contributes +4.1\%/+10.1\% in Rank-1/mAP. LCM also brings notable gains, improving Rank-1/mAP by +3.6\%/+2.5\% on VesselReID and +1.2\%/+1.0\% on Warships-Reid. When both GCM and LCM are combined, MCFormer achieves the best performance: +6.1\%/+8.9\% on VesselReID and +4.4\%/+11.2\% on Warships-Reid. These results clearly show that both GCM and LCM play complementary roles in enhancing performance by modeling global and local inter-image correlations.
\begin{table}
  \centering
  \caption{Effectiveness of main components of MCFormer.}\label{tab:components}
  \begin{tabular*}{0.8\textwidth}{@{\extracolsep{\fill}}c c c c c}
    \hline
    \multirow{2}{*}{\bfseries Method}
      & \multicolumn{2}{c}{\bfseries VesselReID}
      & \multicolumn{2}{c}{\bfseries Warships-ReID} \\
    & \bfseries Rank-1 & \bfseries mAP
    & \bfseries Rank-1 & \bfseries mAP \\
    \hline
    ViT-B/16           & 66.7 & 54.5 & 91.7 & 77.2 \\
    + GCM              & 70.2 & 62.2 & 95.8 & 87.3 \\
    + LCM              & 70.3 & 57.0 & 92.9 & 78.2 \\
    \bfseries MCFormer & \bfseries72.8 & \bfseries63.4
                       & \bfseries96.1 & \bfseries88.4 \\
    \hline
  \end{tabular*}
\end{table}

\begin{figure}[H]
\centering
\includegraphics[width=0.6\textwidth]{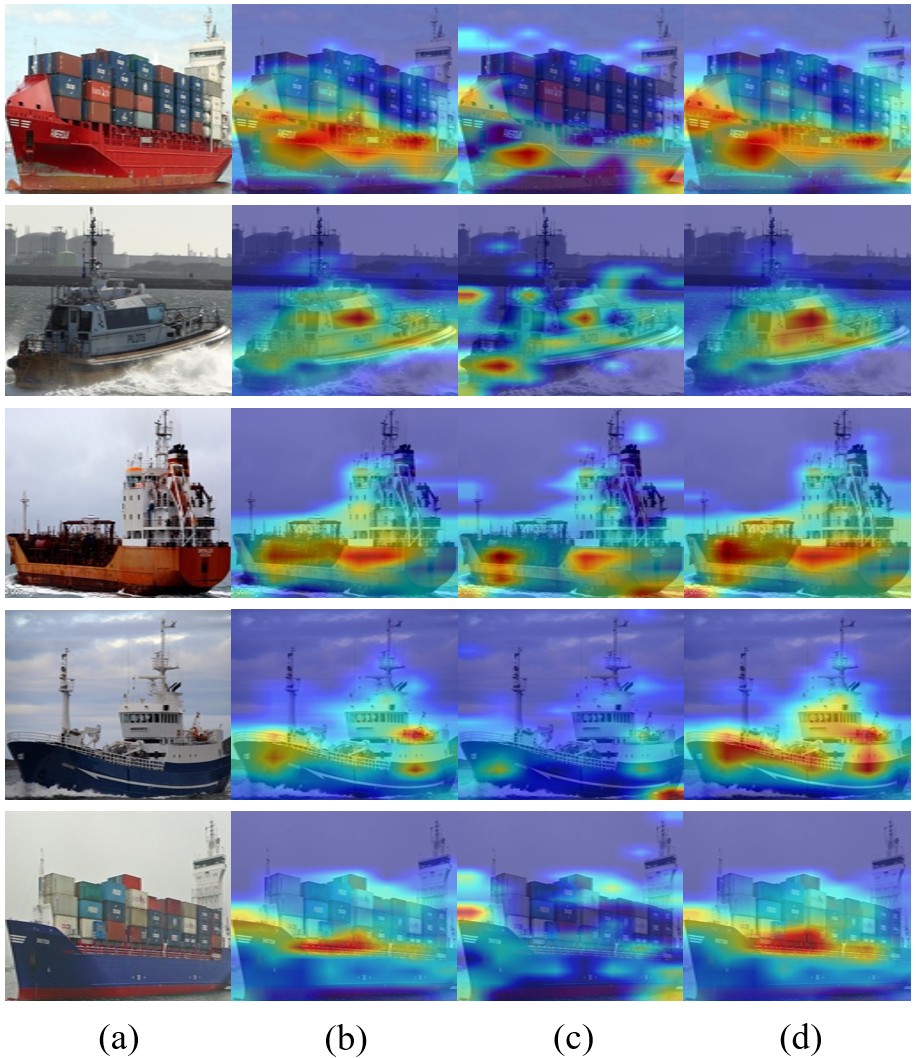}
\caption{Attention maps visualization of the proposed MCFormer under different component settings. (a) Input image of the vessel. (b) Adding GCM only. (c) Adding LCM only. (d) MCFormer.} \label{Attention}
\end{figure}

To further illustrate the contribution of each core module in MCFormer, we present qualitative attention visualizations under different configurations, as shown in Fig.~\ref{Attention}. From the results, we observe that GCM effectively focuses attention on globally salient regions, such as the superstructure and side freeboard, which are often critical for distinguishing vessel identities. In contrast, LCM yields a more dispersed attention pattern across various local parts, reflecting the model's ability to capture complementary fine-grained details. Notably, the full MCFormer combines the strengths of both modules, resulting in a more complete and sharply focused attention distribution that covers both global semantic structure and discriminative local regions. These visualizations support the quantitative gains shown in Table~\ref{tab:components} and validate the complementary effects of GCM and LCM in building a stronger vessel Re-ID model.

\begin{figure}[H]
\centering
\includegraphics[width=0.8\textwidth]{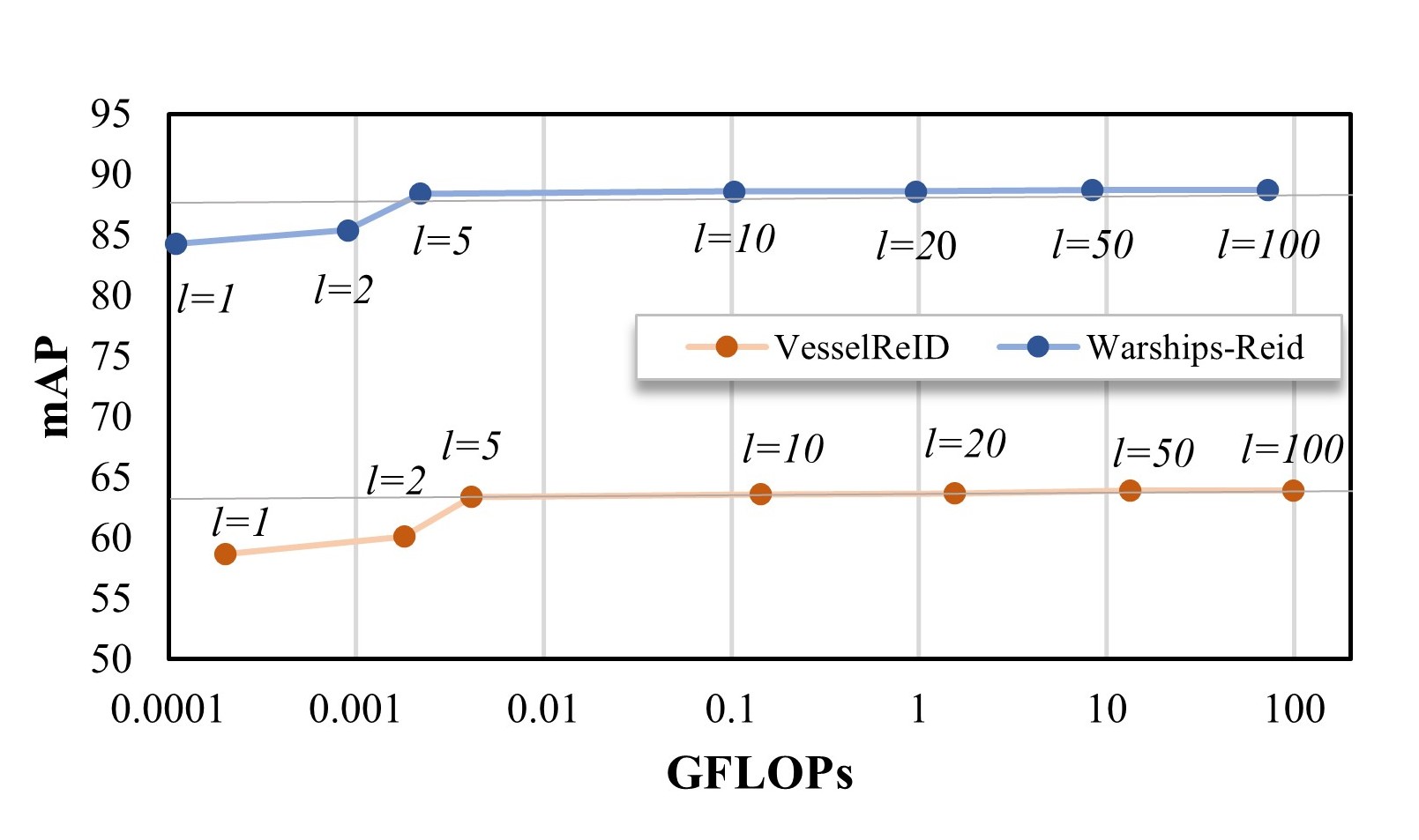}
\caption{This figure illustrates how mAP and GFLOPs change with different values of \(l\). The brown line represents VesselReID, while the blue line represents Warships-ReID.} \label{GCMl}
\end{figure}

\noindent
\textbf{Influence of GCM.}
Having established the effectiveness of GCM, we further analyze its impact by investigating its key hyperparameter, the number of landmarks \(l\). Fig.~\ref{GCMl} illustrates how mAP and GFLOPs vary with different values of \(l\) on the two datasets. Increasing \(l\) from 1 to 5 leads to significant improvements in mAP on both datasets, accompanied by rising computational costs (GFLOPs). However, beyond \(l=5\), mAP saturates while GFLOPs continue to grow, indicating diminishing returns. Based on this trade-off between effectiveness and efficiency, we select \(l=5\) as the optimal setting.

\noindent
\textbf{Influence of LCM.}
In order to effectively capture complementary local correlation information, it is crucial to determine an optimal number of local partitions \(k\). Table~\ref{tab:numparts} demonstrates that segmenting each image into 3 parts achieves the best mAP and Rank-1 performance on the two datasets. If too few partitions are used, different image regions may struggle to capture meaningful local details. Conversely, if too many partitions are used, the clustering process may yield fragments lacking semantic coherence, introducing excessive noise. Thus, setting \(k=3\) provides a good balance between capturing meaningful parts and avoiding fragmentation.
\begin{table}
  \centering
  \caption{Influence of number of parts.}\label{tab:numparts}
  \begin{tabular*}{0.8\textwidth}{@{\extracolsep{\fill}}c c c c c}
    \hline
    \multirow{2}{*}{\bfseries Number of parts}
      & \multicolumn{2}{c}{\bfseries VesselReID}
      & \multicolumn{2}{c}{\bfseries Warships-ReID} \\
    & \bfseries Rank-1 & \bfseries mAP
    & \bfseries Rank-1 & \bfseries mAP \\
    \hline
    2  & 70.9 & 61.6 & 94.8 & 87.1 \\
    \textbf{3}  & \textbf{72.8} & \textbf{63.4} & \textbf{96.1} & \textbf{88.4} \\
    4  & 71.6 & 62.5 & 95.6 & 87.8 \\
    5  & 71.6 & 62.7 & 95.6 & 87.9 \\
    \hline
  \end{tabular*}
\end{table}

\noindent
\textbf{Influence of fusion strategies.}
Having demonstrated the effectiveness of GCM and LCM, it is crucial to select an appropriate global-local feature fusion strategy to enhance the robustness of the final Re-ID representation. Table~\ref{tab:fusion} compares the performance of three fusion strategies on the two datasets: Addition-based fusion, Concatenation-based fusion and Multi-Scale Channel Attention (MCA). As shown, MCA consistently achieves the best Rank-1 and mAP across both datasets, outperforming conventional methods by a clear margin.
\begin{table}
  \centering
  \caption{Comparison of fusion strategies.}\label{tab:fusion}
  \begin{tabular*}{0.8\textwidth}{@{\extracolsep{\fill}}c c c c c}
    \hline
    \multirow{2}{*}{\bfseries Fusion Strategies}
      & \multicolumn{2}{c}{\bfseries VesselReID}
      & \multicolumn{2}{c}{\bfseries Warships-ReID} \\
    & \bfseries Rank-1 & \bfseries mAP
    & \bfseries Rank-1 & \bfseries mAP \\
    \hline
    Addition       & 69.2 & 59.5 & 93.3 & 84.5 \\
    Concatenation  & 70.1 & 61.0 & 95.8 & 86.2 \\
    \bfseries MCA   & \bfseries72.8 & \bfseries63.4
                   & \bfseries96.1 & \bfseries88.4 \\
    \hline
  \end{tabular*}
\end{table}

\section{Conclusion}
In this paper, we propose a novel Multi-Scale Correlation-Aware Transformer Network for vessel re-identification, which explicitly models both global and local correlations across the entire input set to suppress the adverse effects of outlier samples. Specifically, we introduce a Global Correlation Module to capture holistic inter-image dependencies by computing a similarity affinity matrix, and a Local Correlation Module is designed to mine and align local features with contextual similarity. The global and local features that have been respectively correlated are further integrated through a Multi-Scale Channel Attention mechanism. Experimental results show that our method achieves SOTA in three vessel Re-ID benchmarks, validating its effectiveness in capturing latent inter-image relationships among features.
%
%
%
%

\end{document}